\documentclass{article}
\usepackage{times}
\usepackage{amsmath,amssymb}
\usepackage{graphicx}
\usepackage{tabularx}
\usepackage{subfigure} 
\usepackage{algorithm}
\usepackage{algorithmic}
\usepackage{natbib}
\usepackage{hyperref}
\usepackage{cleveref}
\usepackage{xspace}
\usepackage[accepted]{icml2017}
\usepackage{bbold} % conv5, conv6

\usepackage{etoolbox,siunitx} % align table column by the .
\robustify\bfseries % allow bold entries
\usepackage{diagbox} % slash split table
\usepackage{overpic} % (a) and (c)
\usepackage{caption} % captionbox for resultsplate

\usepackage[textwidth=2.cm,textsize=tiny]{todonotes}
\crefname{figure}{Fig.}{Fig.}
\crefname{figure}{Fig.}{Fig.}
\crefname{table}{Table}{Table}

\newcommand{\ninputs}{T_0}

\newcommand{\bn}{\mathbf{n}}

\newcommand{\bp}{\mathbf{p}}
\newcommand{\bq}{\mathbf{q}}
\newcommand{\bx}{\mathbf{x}}
\newcommand{\by}{\mathbf{y}}

\newcommand{\losstwo}{\mathcal{L}}
\newcommand{\lossnormal}{\mathcal{L}_\text{nrm}}
\newcommand{\lossheat}{\mathcal{L}_\text{heat}}

\newcommand{\szero}{S0\xspace} % Folder name: 20170131
\newcommand{\sone}{S1\xspace} % Folder name: 20170202
\newcommand{\stwo}{S2\xspace} % Folder name: 20170216

\newcommand{\SimNet}{\textit{SimNet}\xspace}
\newcommand{\NetOne}{\textit{MechaNet1}\xspace} % V + LSTM + L2 
\newcommand{\NetTwo}{\textit{MechaNet2}\xspace} % T + conv + L2
\newcommand{\NetThree}{\textit{MechaNet3}\xspace} % T + conv + Gauss
\newcommand{\NetFour}{\textit{MechaNet4}\xspace} % T + conv + Soft-max

\newcommand{\anglex}{\theta_{x}}
\newcommand{\angley}{\theta_{y}}
\newcommand{\noentry}{{\text{--}}}

\newcommand{\tensor}{(T)}
\newcommand{\vect}{(V)}

\newcommand{\vnudgeeq}{\vspace{-0.05in}}

\begin{document}

\twocolumn[
\icmltitle{%
%Learning Long-term Predictors of Mechanical Phenomena
Learning A Physical Long-term Predictor}
\icmltitlerunning{Learning A Physical Long-term Predictor}

\begin{icmlauthorlist}
%\icmlauthor{id 594}{}
\icmlauthor{Sebastien Ehrhardt}{ox}
\icmlauthor{Aron Monszpart}{ucl}
\icmlauthor{Niloy J. Mitra}{ucl}
\icmlauthor{Andrea Vedaldi}{ox}
\end{icmlauthorlist}

\icmlaffiliation{ox}{University of Oxford, United Kingdom}
\icmlaffiliation{ucl}{University College London, United Kingdom}

\icmlcorrespondingauthor{Sebastien Ehrhardt}{hyenal@robots.ox.ac.uk}
\icmlcorrespondingauthor{Aron Monszpart }{a.monszpart@cs.ucl.ac.uk}
\icmlcorrespondingauthor{Niloy J. Mitra}{n.mitra@cs.ucl.ac.uk}
\icmlcorrespondingauthor{Andrea Vedaldi}{vedaldi@robots.ox.ac.uk}

%\icmlcorrespondingauthor{Eee Pppp}{ep@eden.co.uk}
\icmlkeywords{physic simulation, machine learning, ICML}
\vskip 0.3in
]
\printAffiliationsAndNotice{}

\begin{abstract} 
Evolution has resulted in highly developed abilities in many natural intelligences to quickly and accurately predict  mechanical phenomena. 
Humans have  successfully developed laws of physics to abstract and model such mechanical phenomena. 
In the context of artificial intelligence, a recent line of work has focused on estimating physical parameters based on sensory data and use them in physical simulators to make long-term predictions. 
In contrast, we investigate the effectiveness of a single neural network for  end-to-end long-term prediction of mechanical phenomena. 
Based on extensive evaluation, we demonstrate that such networks can outperform alternate approaches having even access to ground-truth physical simulators, especially when some physical parameters are unobserved or not known {\em a-priori}. 
Further, our network outputs a distribution of outcomes to capture the inherent uncertainty in the data. 
Our approach demonstrates for the first time the possibility of making actionable long-term predictions from sensor data {\em without} requiring to explicitly model the underlying physical laws.
\end{abstract} 
\section{Introduction}\label{sec:introduction}

Most natural intelligences possess a remarkably accurate understanding of some of the physical properties of the world, as needed to navigate, prey, burrow, or perform any number of other ecologically-motivated activities. In particular, evolutionary pressure has caused most animals to develop the capability to perform fast and accurate predictions of mechanical phenomena. However, the nature of these mental models remains unclear and is being actively investigated~\cite{mentalModel}.

Humans have developed an excellent \emph{formal understanding} of physics; for example, at the level of granularity at which animals operate, mechanics is nearly perfectly described by Newton's laws. However, while these laws are simple, their application to the description of a natural phenomena is anything but trivial. In fact, before such laws can be applied, a physical scenario needs first to be \emph{abstracted} (\textit{e.g.},\ by segmenting the world into rigid objects, describing those by mass volumes, estimating physical parameters). Then, except for the most trivial problems, predictions require the \emph{numerical integration} of very complex systems of equations. It is therefore unclear whether animals would perform mechanical predictions in this manner.

\begin{figure}[t]
    \begin{overpic}[width=\linewidth]{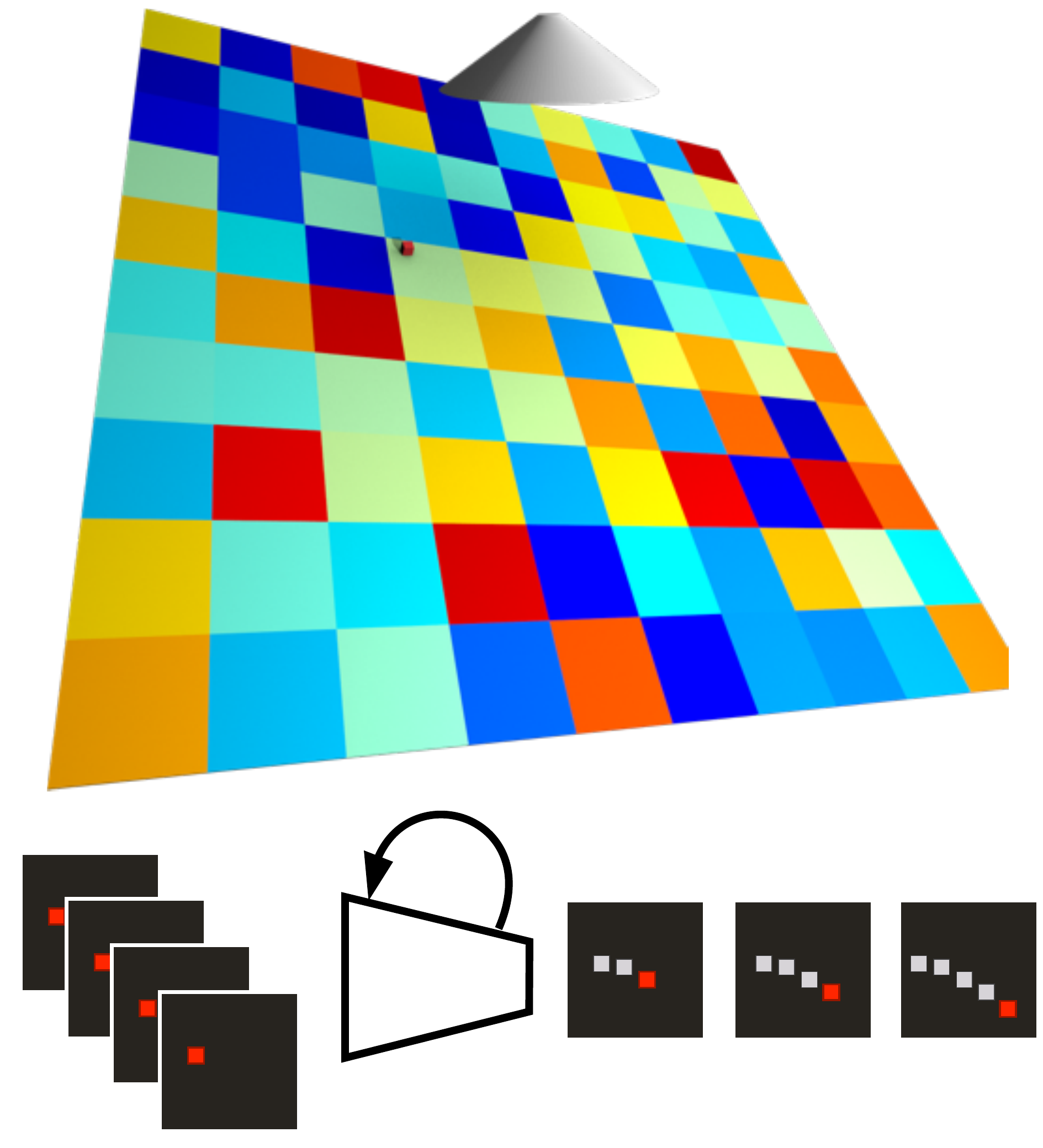}
    \put(60, 93){\small Camera}
    \put(2,-2){\small Input images $t=0\ldots3$}
    %\put(17.5,46){\small $t$=9}
    \put(52,6){\small $t=19$}
    \put(67,6){\small $t=29$}
    \put(81,6){\small $t=39$}
    \put(34,14){\small RNN}
    \linethickness{0.8mm}
    \put(36,79){\color[rgb]{0., 0., 0.}\circle{5}}
    \end{overpic}
    \caption{\textbf{MechaNets.} We consider the problem understanding and extrapolating mechanical phenomena with recurrent deep networks. An orthographic camera looks at a red cube sliding down a black slope with random inclination and heterogeneous friction coefficient (indicated in the top image by the fake coloured tiles). The camera observes the cube for four frames (bottom left) and a recurrent network (bottom middle) predicts the long term motion for up to 40 frames (bottom right). Our goal is to investigate two which extent recurrent networks can develop an internal representation of physics.}    
    % for four frames a sliding cube (   
    % the rendered plane is always a black object (in the inset images, black plane has been replaced by white to improve legibility of this figure). A small red cube rests on the incline and starts sliding due to the effect of gravity. As it slides, it will pass by several different tiles.}
    \label{fig:simulation_setup}
\end{figure}

In this paper, we investigate how an accurate understanding of mechanical phenomena can emerge in artificial systems, mimicking natural intelligence. Inspired by a number of recent works, we look in particular at how deep neural networks can be used to perform mechanical predictions in simple physical scenarios (\cref{fig:simulation_setup}). Among such prior works, by far the most popular approach is to use neural networks \cite{phys101} to \emph{extract from sensory data local predictions of physical parameters}, such as mass, velocity, or acceleration, that are then integrated by an external mechanism such as a \emph{physical simulator} to obtain long term predictions. In other words, these approaches look at how an AI can abstract sensory data in physical parameters, but {\em not} how it can integrate such parameters over longer times. Further, such an approach assumes access to a simulator that accurately abstracts the physical world with appropriate Newtonian equations. Other attempts have also tried to replace the physical engine with a neural network \cite{battaglia2016interaction} but did not really attempt to observe the physical world and deduce properties from it but rather to integrate the physical equations.

% Everyone else is doing iterative updates, we do long term prediction from internal states.

% Position estimation is not done, everyone does velocity estimation and then integrates blindly. Good: more effective, Bad: coordinate frame dependent

By contrast, in this paper we perform end-to-end prediction of mechanical phenomena with a single neural network, implicitly combining prediction and integration of physical parameters from sensory data. In other words, while most other approaches predict instantaneous parameters such as mass and velocity from a few video frames, our model directly performs long-term predictions of physical parameters such as position well beyond the initial observation interval. Thus, as our main contribution, we propose to do so by learning an internal representation of a physical scenario which is induced by the observation of a few images and then is evolved in time by a recurrent architecture.

%While statistical predictors can reduce uncertainty by \emph{learning physical priors}, even the smartest extrapolator cannot be expected to predict the future deterministically.

One of the challenges of extrapolating physical measurements is that the state of a physical system can be determined only up to a certain accuracy, and such uncertainty rapidly accumulates over time. Since no predictor can be expected to deterministically predict the future, predictors are best formulated as estimating \emph{distribution of possible outcomes}. In this manner, predictors can properly account for uncertainty in the physical parameters, approximations in the model, or other limitations. Thus, our second contribution is to allow the neural network to explicitly model uncertainty by predicting distributions over physical parameters rather than their value directly.

In our experiments, we let convolutional neural networks choose their own internal representation of physical laws. A soft prior is that convolutional architectures encourage learning structures that are local and spatially homogeneous, similar to the applicable physical laws that are also local and homogeneous. However, the network is never explicitly told what physical laws are. Our third contribution, therefore, is to look at whether such networks can learn physical properties that generalise beyond regimes observed during training.

The relation of our work to the literature is discussed in~\cref{sec:related}. The detailed structure of the proposed neural networks is given and motivated in~\cref{sec:method}. These networks are extensively evaluated on a large dataset of simulated physical experiments in~\cref{sec:experiments}. A summary of our finding can be found in~\cref{sec:conclusions}.

\section{Related Work}
\label{sec:related}

In this work we address the problem of long-term prediction of object positions in a physical environment without voluntary perturbation with an implicit learning of physical laws. Our work is  closely related to a range of recent works in the machine learning community.

\paragraph{Learning intuitive physics.}
To the best of our knowledge \cite{battaglia2013simulation} was the first approach to tackle intuitive physics with the aim to answer a set of intuitive questions (\textit{e.g.}, will it fall?) using physical simulations. Their simulations, however used a sophisticated physics engine that incorporates prior knowledge about Newtonian physical equations. More recently \cite{Mottaghi_2016_CVPR} also used static images and a graphic rendering engine (Blender) to predict movements and directions of forces from a single RGB image. Motivated by the recent success of deep learning for image processing (\textit{e.g.}, \cite{krizhevsky2012imagenet,he2016deep}) they used a convolutional architecture to understand dynamics and forces acting behind the scenes from a static image and produced a ``most likely motion" rendered from a graphics engine. In a different framework \cite{lerer2016learning} and \cite{li2016visual} also used the power of deep learning to extract an abstract representation of the concept of stability of block towers purely from images. These approaches successfully demonstrated that not only was a network able to accurately predict the stability of the block tower but in addition, it could identify the source of the instability. Other approaches such as \cite{NIPS2016_6113} or \cite{denil2016learning} also attempted to learn intuitive physics of objects through manipulation. None of these approaches did, however, attempt to precisely model the evolution of the physical world.

\paragraph{Learning dynamics.}
Learning the evolution of an object's position also implies to learn about the object's dynamics regardless of any physical equations. While most successful techniques used LSTM-s \cite{Hochreiter:1997:LSM:1246443.1246450}, recent approaches show that propagation can also be done using a single cross-convolution kernel. The idea was further developed in \cite{visualdynamics16} in order to generate a next possible image frame from a single static input image. The concept has been shown to have promising performance regarding longer term predictions on the moving MNIST dataset in \cite{debrabandere16dynamic}. The work of \cite{OndruskaAAAI2016} also shows that an internal hidden state can be propagated through time using a simple deep recurrent architecture. These results motivated us to propagate tensor based state representations instead of a single vector representation using a series of convolutions. In the future we also aim to experiment with approaches inspired by \cite{visualdynamics16}.

\paragraph{Learning physics.} 
Works of \cite{Galileo:NIPS:2015} and its extension \cite{phys101} propose methods to learn physical properties of scenes and objects. However in \cite{Galileo:NIPS:2015} the MCMC sampling based approach assumes a complete knowledge of the physical equations to estimate the correct physical parameters.
In \cite{phys101} deep learning has been used more extensively to replace the MCMC based sampling but this work also employs an explicit encoding and computation of physical laws to regress the output of their tracker.
\cite{Stewart2016LabelFreeSO} also used physical laws to predict the movement of a pillow from unlabelled data though their approach was only applied to a fixed number of frames.

In another related approach \cite{fragkiadaki2015learning} attempted to build an internal representation of the physical world. Using a billiard board with an external simulator they built a network which observing four frames and an applied force, was able to predict the 20 next object velocities. Generalization in this work was made using an LSTM in the intermediate representations. The process can be interpreted as iterative since frame generation is made to provide new inputs to the network. This can also be seen as a regularization process to avoid the internal representation of dynamics to decay over time which is different to our approach in which we try to build a stronger internal representation that will attempt to avoid such decay.

Other research attempted to abstract the physics engine enforcing the laws of physics as neural network models. 
\cite{battaglia2016interaction} and \cite{chang2016compositional} were able to produce accurate estimations of the next state of the world. Although the results look plausible and promising, long term predictions are still an issue in such frameworks. Note, that their process is an iterative one as opposed to ours, which propagates an internal state of the world through time.

\paragraph{Approximate physics with realistic output.}
Other approaches also focused on learning the production of realistic future scenarios (\cite{CNNFluid2016} and \cite{jeong2015data}), or inferring collision parameters from monocular videos~\cite{MonszpartEtAl:SMASH:2016}. In these approaches the authors used physics based losses to produce visually plausible yet erroneous results. They however show promising results and constructed new losses taking into account additional physical parameters other than velocity.

% -------------------------------------------------------------------
\section{Mechanics Networks}\label{sec:method}
% -------------------------------------------------------------------

In this section, we introduce a number of neural network models that can predict the behaviour of a simple mechanical system. We start by describing the physical setup and then we introduce the proposed network architectures.

% -------------------------------------------------------------------
\subsection{Physical setup}\label{sec:phys}
% -------------------------------------------------------------------

The physical setup (\cref{fig:simulation_setup}) consists of a small object sliding down an inclined plane. For notational simplicity, we identify the 3D Euclidean space with the underlying vector space $\mathbb{R}^3$ and denote as $\bp = (p_x,p_y,p_z)\in\mathbb{R}^3$ the coordinates of points as well as of vectors. The plane, which for simplicity passes through the origin, has equation $\pi= \{\bp \in \mathbb{R}^3:\langle \bn, \bp \rangle = 0\}$, where $\bn$ is the plane unit normal vector. In addition to the normal $\bn$, capturing the inclination, the plane has also a Coulomb \emph{friction coefficient} $\rho$, which in the simple case is \emph{homogeneous}, but which can also be a spatially varying quantity $\rho : \pi \rightarrow \mathbb{R}_+$.

We set the camera to be located above the plane, at height $h > 0$, centered at point $(0,0,h)$, and looking downwards along the Z-direction $(0,0,-1)$. The camera axes are aligned to the world axes and the camera projection model is orthographic; in this setting, a world point $\bp$ simply projects to pixel $(p_x,p_y)$ in the image. Note that $h$ does not have an influence on the generated image and can be dropped.

The sliding object is a cube with center of mass $\bq(t) = (q_x(t),q_y(t),q_z(t))$ at time $t$, which projects to pixel $\by(t) = \alpha (q_x(t),q_y(t))+\beta$ in the image. Here we consider $H_i\times W_i = 128 \times 128$ images with pixels of size $\alpha = 1$ and offset $\beta =(-64,-64)$. Initially, the cube is placed at rest on top of the plane at a random location $(q_x^0,q_y^0)$, and then starts to slide under the effect of gravity. The cube motion is also affected by friction. 

An experiment instance is a tuple $\alpha = (q_x^0,q_y^0,\bn,\rho)$, consisting of the values of the initial object position, the plane normal, and the friction coefficient or distribution. The inclination $\bn$ is arbitrary (within limits), such that the object can slide in any direction. These parameters, as well as many other constant parameters described in~\cref{sec:data}, are passed to a physical simulator and rendered to simulate the experiment, resulting in a sequence of color images $\mathcal{X}_T^\alpha=(\bx^\alpha(0),\bx^\alpha(1),\dots,\bx^\alpha(T-1))$. The simulator also produces the ground-truth center of mass projections $\mathcal{Y}_T^\alpha = (\by^\alpha(0),\dots,\by^\alpha(T-1))$. Note that, for the purpose of learning predictors, physics needs not to be specified further; while in fact a complete set of parameters are required to run the physical simulation, the predictor learns automatically to extract the required information from the observed images.

% -------------------------------------------------------------------
\subsection{Neural network architectures}\label{sec:nets}
% -------------------------------------------------------------------

\begin{figure}[t]
    \centering
    \begin{overpic}[width=\linewidth]{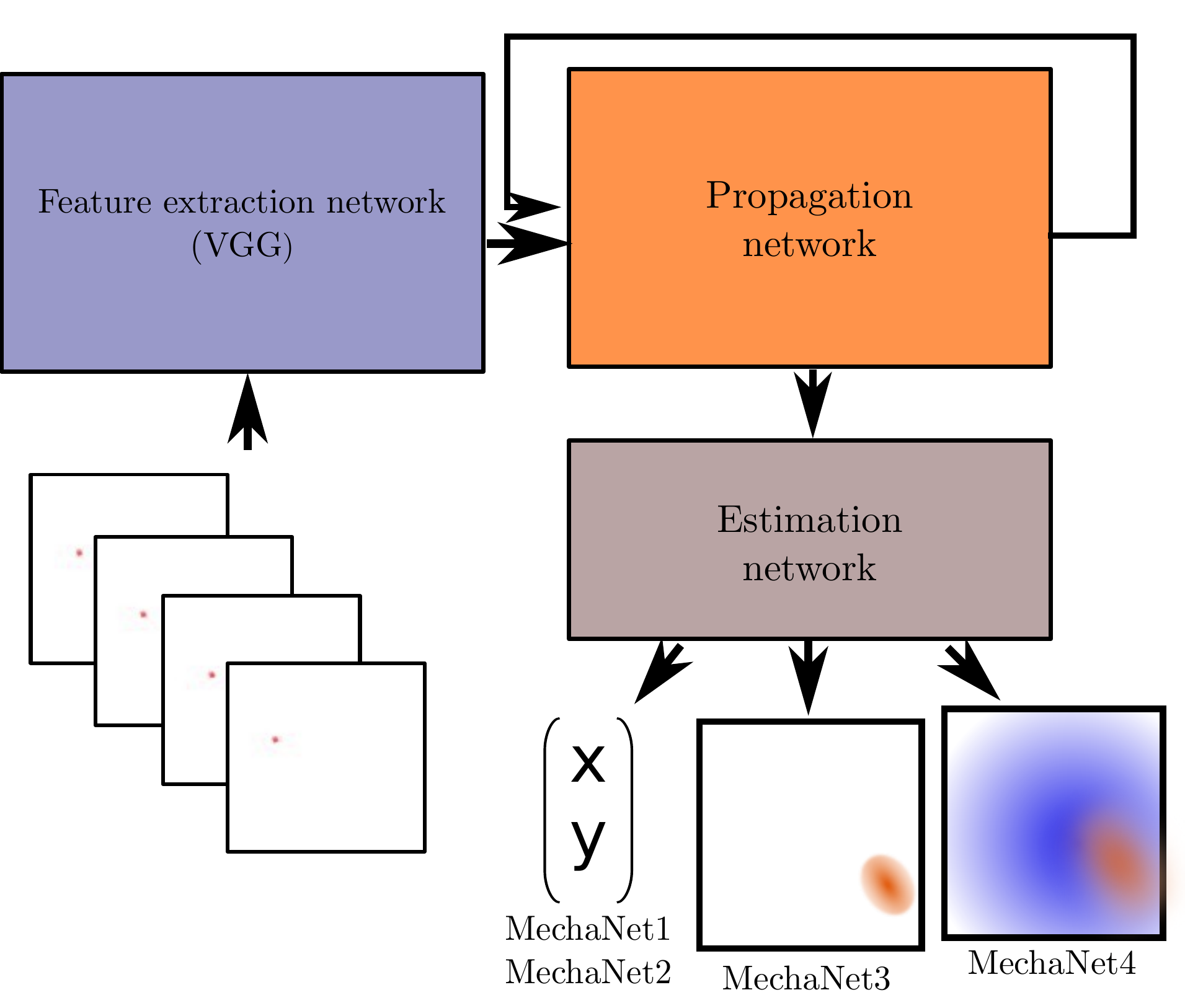}
    \put(1,6){\small Input images $t=0\ldots3$}
    \end{overpic}
    \caption{Overview of our proposed pipeline. The first four images of a sequence first pass through a partially pre-trained feature extraction network to build the concept of physical state. It then recursively passes through a propagation layer to produce long-term predictions about the future positions of the objects. Extrapolation requires us to handle the notion of uncertainty, which is why \NetFour\ performs the best under changing physical conditions in \mbox{Scenario~\stwo}, see \cref{tab:results}. }
    \label{fig:over}
\end{figure}

We focus on long-term predictors $\Phi :\mathcal{X}_{T_0} \mapsto \mathcal{Y}_T$ that take as input the first $T_0=4$ frames $\mathcal{X}_{T_0}$ of a video sequence $\mathcal{X}_T$ and produce as output a long-term estimate $\mathcal{Y}_T$ of the location of the object's center of mass at times $t=0,1,\dots, T$, where $T \gg T_0$.

Our method comprises \textbf{three building blocks} (\cref{fig:over}): a feature extractor, a propagation network and an estimation network. The core of our model is the internal representation of the physics, initialized by the feature extractor, updated by the propagation module, and decoded by the estimation module. We compare two representation types: a \emph{vector representation}, in which each frame is encoded as $C$-dimensional vector (or $1 \times 1 \times C$ tensor), and a $H \times W \times C$ \emph{tensor representation}. The importance of this difference is that the vector representation is spatially \emph{concentrated}, whereas the tensor representation is spatially distributed.

Next, the three modules are discussed in detail.

\paragraph{(i)~Feature extraction network.}
The predictor $\mathcal{Y}_T = \Phi(\mathcal{X}_{T_0})$ starts by extracting information from $T_0$ video frames. Similarly to~\cite{fragkiadaki2015learning}, the RGB channels of the images are concatenated in a single $H_i \times W_i \times 3T_0$ tensor and this is processed by a convolutional neural network $\phi_\text{init}$, obtaining a $\phi_\text{init}(\bx_0, \ldots \bx_{T_0-1}) \in \mathbb{R}^{H\times W\times C}$ tensor output. These features serve as the internal physical representation of our network that is propagated through time. Inspired by \cite{fragkiadaki2015learning}, we start from the VGG16 network pre-trained on \mbox{ImageNet~\cite{Simonyan15}}. The network is cut and the last layer adapted as needed. In particular, starting from a $(H_i,W_i)=(128,128,3)$ image, the vector representation uses the $(H,W,C)=(1,1,128)$ dimensional output of layer \texttt{fc6}, and the tensor representation uses the $(8,8,512)$ dimensional output of \texttt{conv5} instead. All feature extraction layers are frozen in training, except the new layer \texttt{fc6} and \texttt{conv1}, whose shape changes.

\paragraph{(ii)~Propagation network.} 
The internal representation initialized by the feature extractor is evolved through time by the propagation network $F : \mathbb{R}^{H\times W\times C} \rightarrow \mathbb{R}^{H\times W\times C}$. Formally, the internal state $S_t$ is initialized as $S_0 = \phi_\text{init}(\mathcal{X}_{T_0})$ and updated by iteratively applying $F$ as $S_{t+1} = F(S_t) = F^{t}(\phi_\text{init}(\mathcal{X}_{T_0}))$ for $t\geq 1$ (note that there is an index shift between state and time, so that $S_t$ predicts the objct position at time $t+T_0-1$). For the vector representation, the propagation network is an LSTM with 128 hidden units. Since there are no more observations after $T_0$, the LSTM input at time $t$ is set to the internal state of the LSTM at the previous time. This is  similar in approach to \cite{cho-al-emnlp14}, although our output is directly fed to the network without re-embedding. For the distributed representation, we use a simple chain of two convolution layers (with $256$ and $512$ filters respectively, of size $3\times3$, stride 1, and padding 1) interleaved by a ReLU layer. When using \textit{discrete probability map}, the representation $S_t$ is normalised channel-wise in $L^2$ norm after each update in order to avoid the decay of intermediate propagation layers.

\paragraph{(iii)~Estimation network.}
In the simplest instance, the network predictor estimates directly the values $\widehat{\mathcal{Y}}_T=(\hat \by(0),\dots, \hat \by(T-1))$ of the object's center of mass $\widehat{\by}(t)\in\mathbb{R}^2$ during the sequence. The learning loss is simply the average squared distance between predicted and ground-truth locations:
\vnudgeeq
\[
   \losstwo(\widehat{\mathcal{Y}}_T,\mathcal{Y}_T)
   =\frac{1}{T} \sum_{t=0}^{T-1} \| \hat \by(t) - \by(t) \|^2.
\vnudgeeq
\]
As discussed above, however, it is preferable to predict the \emph{uncertainty} of the estimate as well. While in some cases this cannot improve accuracy directly (i.e in the bivariate gaussian case), it is interesting to see if a network is able to develop an internal sense of prediction errors. Further, probabilistic modelling may help the network discount difficult-to-predict points during training, which may otherwise work as outliers negatively affecting training.

We propose to do so in two ways. In the first approach, we predict the mean and variance $\widehat{\mathcal{Y}}_T=(\mu(t),\Sigma(t);t=0,\dots,T-1)$ of a bivariate Gaussian distribution $\mathcal{N}(\cdot;\mu,\Sigma)$. The loss is the negative log-likelihood of the measured object locations:
\vnudgeeq
\[
 \lossnormal(\widehat{\mathcal{Y}}_T,\mathcal{Y}_T)
   = - \frac{1}{T} \sum_{t=0}^{T-1} \log \mathcal{N}(\by(t); \mu(t),\Sigma(t)).
\vnudgeeq
\]
In practice, the neural network estimates the two dimensional vector $\mu(t)$ as well as a three dimensional vector $\lambda_1(t), \lambda_2(t), \theta(t)$ with the first two  being the eigenvalues of $\Sigma(t)$, and the third entry being the angle of the rotation matrix in the decomposition $\Sigma(t) = R(-\theta(t)) \begin{bmatrix}
    \lambda_1(t) & 0\\
         0 & \lambda_2(t) 
\end{bmatrix}R(\theta(t))$. In order to ensure numerical stability, eigenvalues are constrained to be in the range $[0.01 \ldots 100]$ by setting them as the output of a scaled and translated sigmoid $\lambda_i(t) = \sigma_{\lambda,\alpha}(\beta_i(t))$, where $\sigma_{\lambda,\alpha}(z) = \lambda/(1 + \exp(-z)) + \alpha$.
%
%To summarise, the network outputs $\mu$(t), $\theta$(t), $\beta_1$(t), $\beta_2$(t) and obtains $\Sigma(t)$ as $R(-\theta(t)) \begin{bmatrix}
%    \sigma_{\lambda,\alpha}(\beta_1(t)) & 0\\
%          0 & \sigma_{\lambda,\alpha}(\beta_2(t)) 
% \end{bmatrix}R(\theta(t))$.
%
For more details regarding the training procedure of this model please see section~\ref{sec:implem}.

In the second approach, we predict \emph{discrete probability maps} $\mathcal{Y}_T=(p(0),\dots,p(T-1))$, where $p(t) \in \mathbb{R}^{H_i\times W_i}$ and $p(t)_{ij}$ is the probability that the object's center of mass is contained in a $1 \times 1$ square centered at location $(j-W_i/2,i-H_i/2)$. Similar to the Gaussian loss, we use the negative log-likelihood of the ground-truth observations as loss:
\vnudgeeq
\vnudgeeq
\[
 \lossheat(\widehat{\mathcal{Y}}_T,\mathcal{Y}_T)
   = 2 \log \delta - \frac{1}{T} \sum_{t=0}^{T-1} \log p(t)_{\left\lfloor\by(t)\right\rceil},
\]
where $\lfloor \cdot \rceil$ is the rounding operator and $\delta=1$ is the sampling step.\footnote{The correction $2\log \delta$ results in the log-likelihood value of the piecewise-constant continuous distribution corresponding to the discrete one and makes likelihood values comparable for different step sizes $\delta$, as well as comparable to the Gaussian log-likelihood.} The probability maps $p(t)$ sum to one and are obtained by applying the softmax operator to a tensor $A(t)\in\mathbb{R}^{H_i \times W_i}$ estimated by the neural network:
\vnudgeeq
\[
  p(t)_{ij} = 
  \frac{e^{A(t)_{ij}}}{\sum_{mn}e^{A(t)_{mn}}}.
\vnudgeeq
\]
All the output predictions at time $t+T_0-1$ are extracted from the internal state $S_t$ by a single layer $L(S_t)$. The layer $L$ is linear and fully-connected layer for $\losstwo$ and $\lossnormal$, and a deconvolutional layer similar to \cite{Long_2015_CVPR} in the case of $\lossheat$. Outputs at times $t=0,1,\dots,T_0$ are all predicted from $S_0$ using an analogous but independently-trained fully-connected layer $L'$.
%Except for the probability map $L$ will be a fully-connected layer otherwise a deconvolution layer similar to \cite{Long_2015_CVPR}. 
Overall, the output of the predictor is given by:
\begin{align*}
   \Phi(\mathcal{X}_{T_0}) = (L'(S_{0})_1, &\ldots , L'(S_{0})_{T_0},
   \\& L(S_{1}),  \ldots, L(S_{T-T_0-1})). 
\end{align*}

% -------------------------------------------------------------------
\section{Experiments}\label{sec:experiments}
% -------------------------------------------------------------------
%\aron{4.1 Scenarios, 4.2 Networks, 4.3 Implementation details}

% -------------------------------------------------------------------
\subsection{Data generation}\label{sec:data}
% -------------------------------------------------------------------

Experiments consider three variants of the physical setup described in~\cref{sec:phys}, called Scenarios \szero, \sone, and \stwo. Different scenarios sample experiments $\alpha = (q_x^0,q_y^0,\bn,\rho)$ of increasing difficulty. The parameters of each scenario are summarised in~\cref{tab:datasets} and described next. The plane normal $\bn$ was obtained by rotating the $Z$ axis around the $X$ and $Y$ axis by random angles $\anglex, \angley$ (Scenario \szero uses a fixed inclination). For Scenarios \sone and \stwo, the Coulomb friction coefficient $\rho$ of the plane is homogeneous and sampled uniformly at random. For Scenario \stwo, the plane is split in $10\times10$ patches, each with a random friction coefficient sampled independently. The friction upper bound was chosen so that the object always slides along the slope.

\begin{table}[h!]
\footnotesize
\centering
\setlength{\tabcolsep}{2pt}
\begin{tabular}{|c|c|c|}
\hline
Scenario & $\bn$ rotation & $\rho$ \\
\hline
\szero & $\anglex=0,~\angley=\frac{\pi}{6}$ & $\rho \in \mathcal{U}\left(10^{-4},10^{-1}\right)$ \\
\sone  & $\anglex,\angley\in\mathcal{U}\left(-\frac{\pi}{6}, \frac{\pi}{6}\right)$ & $\rho \in \mathcal{U}\left(10^{-4},10^{-1}\right)$ \\
\stwo  & $\anglex,\angley\in\mathcal{U}\left(-\frac{\pi}{6}, \frac{\pi}{6}\right)$ & $\forall_{i,j=1\ldots {10}}~\rho_{i,j} \in \mathcal{U}\left(0.5,5\right)$ \\
\hline
\end{tabular}
\caption{\textbf{Data generation parameters.}
%We vary the rotation of the plane along at most two axes, and randomize the friction properties of the plane in order to ensure, that our internal physics representation generalizes over different influences of gravity.
}\label{tab:datasets}
\end{table}

%\sisetup{input-ignore={,},input-decimal-markers={.},group-separator={,}}
\begin{table*}
%\begin{minipage}[b]{0.68\linewidth}x
\footnotesize
\setlength{\tabcolsep}{1pt}
%\begin{tabular}{|c|ccc|cccc|cccc|cccc|}
\sisetup{detect-weight=true,detect-inline-weight=math}
\begin{tabular}{|c|ccc|*2S[table-format=-2.2]|*2S[table-format=-2.2]|*2S[table-format=-2.2]|}
\hline
& & & &
\multicolumn{2}{c|}{\szero} &
\multicolumn{2}{c|}{\sone} &
\multicolumn{2}{c|}{\stwo}
\\
Method & Feat.\ & Prop.\ & T.~Obj.\ & 
\multicolumn{2}{c|}{$L^2$ (ln perplexity)} &
\multicolumn{2}{c|}{$L^2$ (ln perplexity)} &  
\multicolumn{2}{c|}{$L^2$ (ln perplexity)} \\
& & & & {20} & {40} & {20} & {40} & {20} & {40} 
\\
\hline
\text{Linear}    & \noentry & \noentry & \noentry & 12.62 & 49.07 & 11.81 & 42.58 &  11.86 & 42.37 \\
Quadratic & \noentry & \noentry & \noentry & 6.67 & 37.95 & 8.21 & 46.34 & 8.35  & 47.29 \\
\SimNet & \vect & \noentry & \noentry  & 1.21 & \bfseries 1.89 & 1.93 & \bfseries 5.16  & 5.67 & 24.69 \\
\NetOne   & \vect & LSTM & L2  & \bfseries 0.31 & 25.37 & \bfseries 1.52 & 30.84 &  \noentry & \noentry \\
\NetTwo   & \tensor & conv & L2  & 0.77 & 10.68 & 1.91 & 34.46 & \bfseries 1.95 & 13.24  \\
\NetThree & \tensor & conv & GL & 0.55 & 15.74  & 2.26 & 26.04 &  \noentry & \noentry  \\
& & & &   {(0.97)} & {(385)} &  {(4.08)} & {(36)}&  \noentry &   \noentry
\\
\NetFour  & \tensor & conv & SM& 0.56 & 24.59 & 2.13 & 19.54  & 3.62 & \bfseries 9.55   \\
& & & &   {\bfseries (0.49)} & {\bfseries (0.96)} &    {\bfseries (3.19)} & {\bfseries (17.2)} & {(5.24)}  & {(11.47)} \\
\hline
\end{tabular}\hfill%
%\end{minipage}%
\begin{minipage}[c]{0.32\linewidth}%
\includegraphics[width=\textwidth,trim=0 30pt 0 20pt]{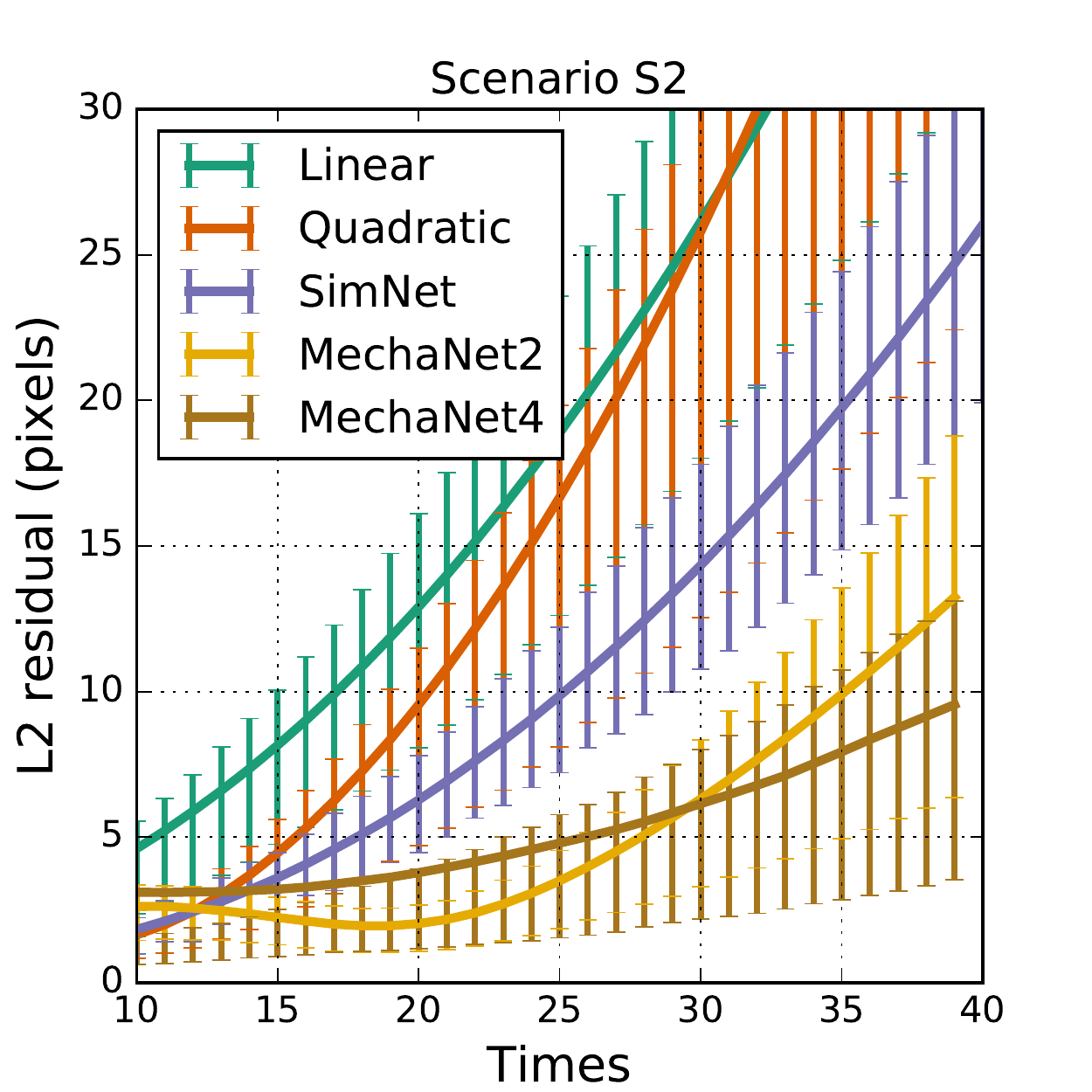}
\end{minipage}
\caption{\textbf{Long term predictions.} \vect\ and \tensor\ refers to the dimensionality of the internal state representations (vector and tensor respectively). We expect \tensor\ to maintain a 2D spatial model which leads to higher accuracy. The \SimNet\ and all \textit{MechaNet} models observed the $\ninputs=4$ first frames as input. All networks have been trained to predict the $T=20$ first positions, except in \mbox{Scenario~\stwo}, where \NetOne\ and \NetFour\ have been trained to predict $T=30$ frames in order to experience enough variation in the underlying physical conditions, \textit{i.e.}, changing friction. Perplexity ($\log_e$ values shown in the table) is defined as $2^{-\mathbb{E}[\log_2(p(x))]}$ where $p$ is the estimated posterior distribution. \emph{Right:} error evolution on experiment S2 for all time steps up to 40. Error bars denote $25^{th}$ and $75^{th}$ percentiles of the $L^2$ loss in pixels.}
\label{tab:results}
\end{table*}

The plane was rendered as a black object so that no static visual cues allow deducing any of the physical parameters except the initial position of the cube; instead the predictor has to approximate physics as needed by observing the motion of the object during the first $\ninputs=4$ frames of each experiment.

Each experiment was run for at most 240 frames, or terminated early if the object left the field of view. In order to observe enough movement in each recorded sequence, the first 30 video frames of each video were removed, and the rest of the video was sub-sampled by a factor of 3. In practice, most experiments consist of 40-50 images. 

The dataset contains 12,500 experiments for each Scenario, 70\% of which are used for training,  15\% for validation, and 15\% for test.

\paragraph{Implementation details.}
The object's starting position is initialized randomly using rejection sampling in such a way that $(q_x^0,q_y^0)$ falls in the slope quadrant that contains the largest visible $h$ coordinate. %\textit{E.g.}, for $\anglex < 0$ we only accept the normalized screen space coordinates $0.0 < q_x^0 < 0.49$ and $0.51 < q_x^0 < 1.0$ otherwise, and similarly for $\angley$.
This procedure generates samples that have most of their trajectory visible to the camera.

Rendering and physical simulation use Blender 2.77's OpenGL renderer and the Blender Game Engine (relying on Bullet 2 as physics engine) respectively. The object is a cube of side $0.1^3$ Blender units with \mbox{mass = 1}. The simulation parameters are: \mbox{max physics steps = 5}, \mbox{physics substeps = 3}, \mbox{max logic steps = 5}, \mbox{FPS = 120}. Rendering used white environment lighting (energy = 0.7) and no other light sources. The object color was set to Lambertian red (RGB: $0.8,~0.04,~0.04$) with no specular component. The slope is completely black, covering the whole field of view. The output images were stored as $128\times128$ color JPG files. See \cref{fig:simulation_setup} for an example initial setting from \mbox{Scenario~\stwo}. 

% -------------------------------------------------------------------
\subsection{Baseline predictors}\label{sec:networks}
% -------------------------------------------------------------------
\paragraph{Least squares fit.} We compare the performance of our methods to two simple least squares baselines: Linear and Quadratic. In both cases we fit two least squares polynomials to the \emph{estimated} screen-space coordinates of the first $T=10$ frames. The polynomials are of first and second degree(s), respectively. We estimate the object's position in this case by using the maximum location of the red channel of the input image. Note, that being able to observe the first 10 frames is a large advantage compared to the networks, which only see the first $\ninputs=4$ frames.
%either first order (linear) or second order (quadratic) polynomial with least squares method on each coordinates estimated on the 10 first object positions. 
%We also aim to compare to other methods, such as \cite{battaglia2016interaction} or \cite{fragkiadaki2015learning} in the future \todo{Aron: I don't think we should say this last sentence}.

\paragraph{Physics simulator.} The \SimNet baseline is used to evaluate the long term prediction ability of a neural network that has access to an explicit physics simulator, in a manner analogous to the work of~\cite{Galileo:NIPS:2015}.
%compare to which aims to use the laws of physics encoded in a physical engine. 
Similarly to the other networks, \SimNet observes the first $\ninputs$ images and aims to regress the physical parameters necessary to predict the trajectory of the object using the physics engine. The simulator is assumed to have access to a perfect model of the underlying physical laws. The regression architecture constitutes of the vector based feature extraction network described in \mbox{Section~\ref{sec:nets}} with an extra fully-connected layer on top to regress physical parameters.
The network is trained with an $L^2$ loss to infer the current slope rotation angles and friction coefficient ($\anglex, \angley, \rho$), the object's position at the observed frames ($t_0,\ldots,\ninputs-1$), and its final velocity at frame $\ninputs-1$.
%To this end we used the feature extraction network of \ref{sec:nets} with output $1 \times 1 \times 128$ and use these features to estimate the following physical parameters:

We input the regressed parameters to the \emph{same} physics simulation system that generated the dataset (\mbox{Section~\ref{sec:data}}) and run the simulation to predict the following object positions $\ninputs \ldots T$. Note that, since the simulator used by the network is the same as the one used to generate the data, this network is given a significant advantage over the other models.
%to generate the next object positions at time $T_0+1 \ldots T$ and compare with our ground truth.

%% This image should be changed
%\begin{figure}[h!]
%    \includegraphics[width=\linewidth]{images/simulator_exp}
%    \caption{Simulator experiment \aron{Do we need this figure?}}
%\end{figure}

\begin{figure*}[t!]
\newcommand{\incx}[1]{\includegraphics[width=0.33\linewidth,trim=20px 40px 30px 40px]{#1}}
\incx{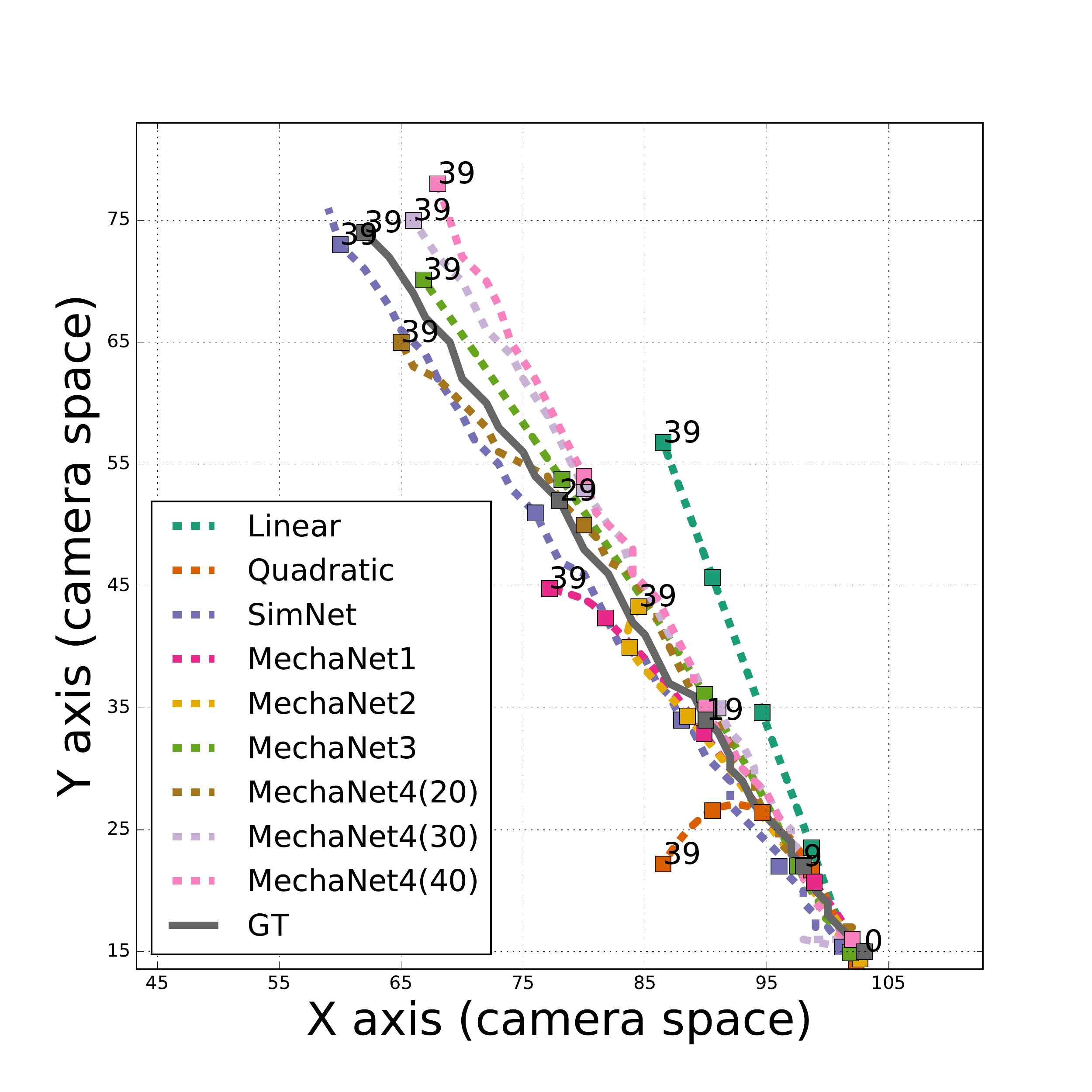}\hfill%
\incx{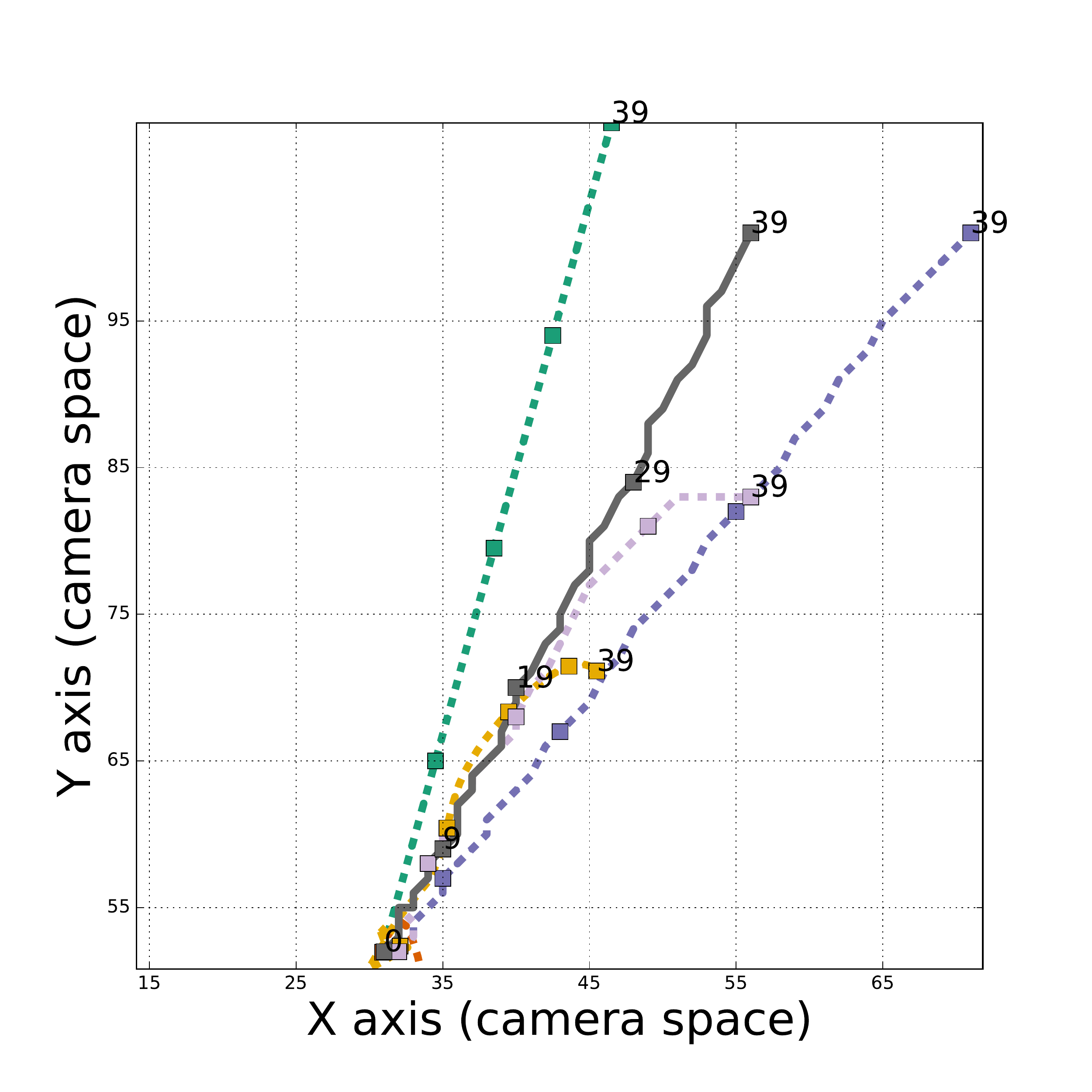}\hfill%
\incx{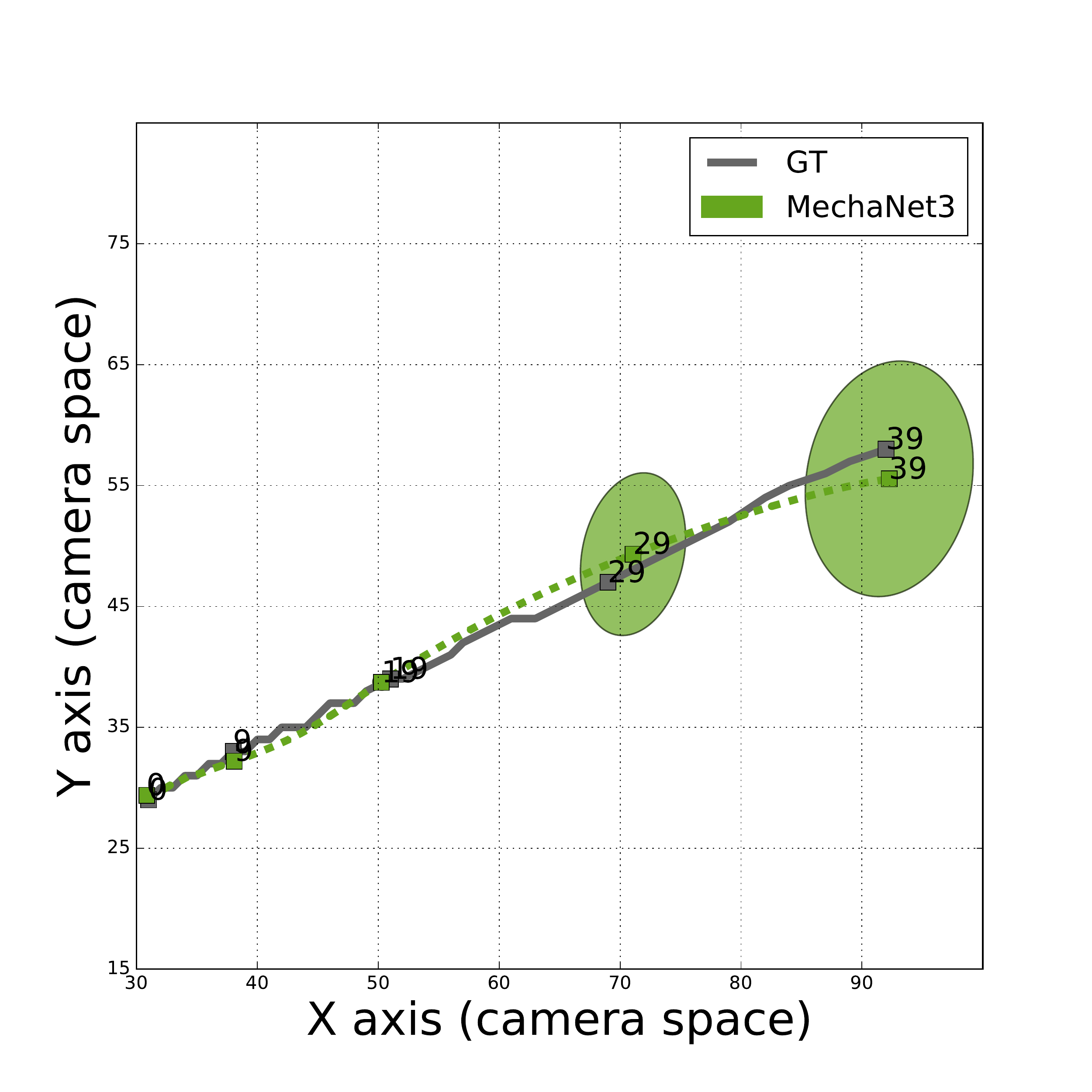}\\
\mbox{}\hfill(a)\hfill\hfill(b)\hfill\hfill(c)\hfill\mbox{}
\vspace{-1em}
\caption{\textbf{Qualitative results.} (a) Predictions of various networks in an example drawn from Scenario S1. The number of frames used to train each network is indicated in parenthesis; if omitted, it defaults to 20. (b) The same but for Scenario S2 and focusing on linear and quadratic fits, \SimNet, \NetTwo and \NetFour (30). (c) Example of uncertainty prediction of \NetThree in Scenario S1.}\label{fig:qualitative}
\end{figure*}

\begin{table}[b!]
\centering
\footnotesize
%\begin{tabular}{|c|cccc|cc|}
\sisetup{detect-weight=true,detect-inline-weight=math}
\begin{tabular}{|c|*4S[table-format=-2.]|*2S[table-format=2.]|}
\hline
{} & {MN4} & {MN4} & {MN4} & {MN4} & {QD} & {SN} \\
\hline
%\diagbox{test}{train} 
 & {10} & {20} & {30} & {40} & {--} & {--}
\\
\hline
10  & 1.18 & 1.60 & \bfseries 0.83 & 1.00 & 1.45 & 1.26 \\
20  & 11.79 & 2.13 & \bfseries 1.36 & 1.38 & 8.21 & 1.93 \\
30  & 28.04 & 6.91 & 2.71 & \bfseries 2.32 & 23.33 & 3.23 \\
40  & 48.65 & 19.54 & 8.96 & \bfseries 4.00 & 46.34 & 5.16 \\
\hline
\end{tabular}
\caption{\textbf{Generalization capabilities.} We compare predictions obtained at different times in \mbox{Scenario~\sone} from four versions of the \NetFour~(MN4) model that have been trained to predict the first $t=10, 20, 30, \text{and }40$ frames (rows). The train and test inputs always consists of the first $\ninputs=4$ images. For comparison, we also show the prediction accuracy of the quadratic baseline~(QD) (fit to the first 10 inputs) and \SimNet~(SN). The numbers represent the average, $L^2$ loss measured in pixels (input image size: $128\times128$).}
\label{tab:generalization}
\end{table}

% -------------------------------------------------------------------
\subsection{MechaNets}\label{sec:implem}
% -------------------------------------------------------------------
%\footnote{\url{http://www.cs.toronto.edu/~tijmen/csc321/slides/lecture_slides_lec6.pdf}}

Experiments consider four different variants of the mechanical prediction networks (\emph{MechaNet1} to 4 for short). \NetOne and \NetTwo are trained to optimise the $\losstwo$; \NetOne uses the LSTM propagation network and the spatially concentrated internal representation, whereas \NetTwo uses the simpler convolutional propagation network but the distributed representation. \NetThree and \NetFour are similar to \NetTwo, but they use probabilist predictors, using the Gaussian and probability map outputs, respectively. The four variants are summarized in~\cref{tab:results}.

\paragraph{Implementation details.} Network weights are initialized by sampling from a Gaussian distribution. Training uses a batch size of 50 using the first 10 to 40 frames of each video sequence using RMSProp~\cite{Tieleman2012}. Training is halted when there is no improvement of the $L^2$ loss after 40 consecutive epochs; 1,000 epochs were found to be usually sufficient for convergence. 

Since during the initial phases of training the network is very uncertain, the model using the Gaussian log-likelihood loss $\mathcal{L}_\text{nrm}$ was found to get stuck on solutions with very high variance $\Sigma(t)$; to solve this issue, the regularizer $\lambda \sum_t \det \Sigma(t)$ was added to the loss, setting $\lambda=10$ for the first few epochs and then lowering it to $\lambda = 0$ when the value of the determinant stablized under 100 on average (this variance is comparable to the image size).

In all our experiments we used Tensorflow \cite{tensorflow2015-whitepaper} r0.12 on a single NVIDIA Titan X GPU. 
% -------------------------------------------------------------------
%\section{Results}\label{sec:result}
% -------------------------------------------------------------------

\begin{figure*}[t!]
\includegraphics[width=1.0\linewidth,trim=480 0 0 0, clip]{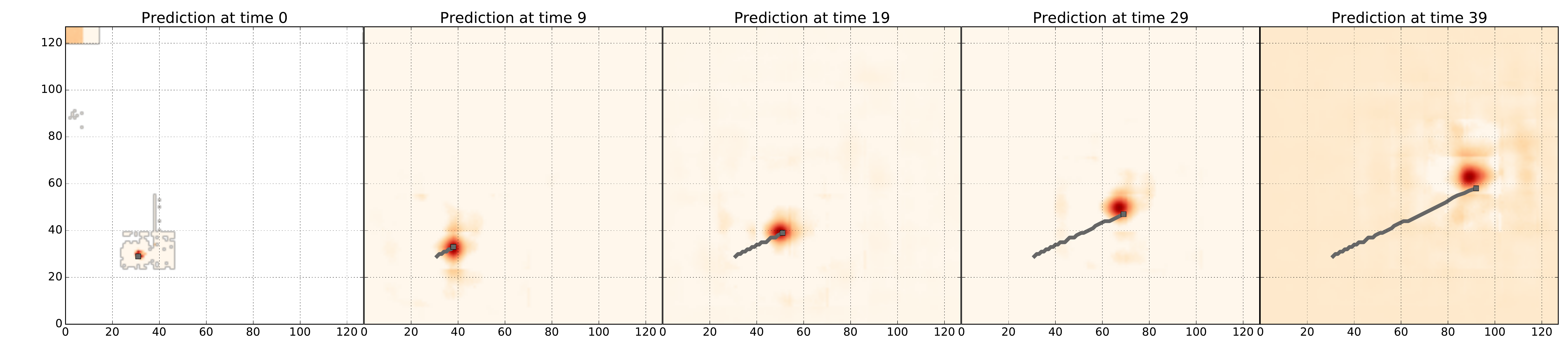}
\caption{\textbf{Uncertainty prediction using probability maps.} The figure shows the output of \NetFour (40) on one example sequence in Scenario S1.}\label{fig:heat}
\end{figure*}

% -------------------------------------------------------------------
\subsection{Results}\label{sec:quant_eval}
% -------------------------------------------------------------------

\paragraph{Long term predictions.} \Cref{tab:results} and \cref{fig:qualitative} compare the baseline predictors and the four MechaNets on the task of long term prediction of the object trajectory. We call this ``long term'' in the sense that all methods observe only the first $T_0=4$ frames of a video (except the linear and quadratic extrapolators which observe the first 10 frames instead), to then extrapolate the trajectory to 40 time steps.

All networks can be used to perform arbitrary long predictions; the table, in particular, reports the the average $L^2$ prediction errors at time $T_\text{test}=20$ and $40$. However, models in this table are only shown the first $T_\text{train}=20$ frames of each video during training.

Consider first the prediction results for $T_\text{test}=T_\text{train}=20$. All networks perform considerably better than the linear and quadratic extrapolators in all scenarios, with error rates 5-40$\times$ smaller. As expected, Scenario S1 and S2 are harder than Scenario S0, which uses a fixed slope and homogeneous friction, but the network prediction errors are still small, in the order of 1-2 pixels. All networks perform similarly well, particularly in Scenarios S1, with a slight advantage for the LSTM-based propagation networks. \mbox{\SimNet} is very competitive, as may be expected given that it uses the ground-truth physics engine for integration. However, in Scenario S2 this method does not work well since the variable friction distribution is not observable from the first $T_0=4$ frames of a video;  \NetTwo and \NetFour, which can better learn such effects, can account for such uncertainty and significantly outperform \SimNet.

%on the other hand, the Gaussian (\NetThree) and softmax (\NetFour) networks predict the output uncertainty as well, as discussed below.

Results are different for predictions at time $T_\text{test}=40 \gg T_\text{train}$. All networks still outperform the extrapolators, but in Scenarios S0 and S1 \SimNet performs better than the other networks: by having access to the physics simulator, generalization is not an issue. On the other hand, this experiment shows that the deep networks have a limited capacity to generalize physics beyond the regimes observed during training. Among such networks, the ones modelling uncertainty (\NetThree and \NetFour) are able to generalize better. Scenario S2 still breaks the assumptions made by \SimNet, and the other networks outperform it.

\paragraph{Generalization.} The issue of generalization is explored in detail in~\cref{tab:generalization}, focusing on \NetFour that exhibits the best generalization capabilities. The table reports prediction errors at $T=10,20,30,40$ for networks trained with video sequence of length $T=10,20,30,40$ respectively. Recall that predictors always observe only the first $T_0=4$ frames of each sequence; the only change is to allow the training loss to assess the predictors' performance on progressively longer videos during training.

As expected, training on longer sequences dramatically improves the accuracy of longer term predictions, but also the shorter term ones. Training on the full sequences, in particular, performs $\sim20\%$ better than \SimNet. This confirms that, while deep networks are able to learn physical rules accurately for the range of physical experiences observed during training, they do not necessarily learn rules that generalize as readily as conventional physical laws.

\paragraph{Predicting uncertainty.} \NetThree and \NetFour predict a posterior distribution of possible object locations, using a Gaussian and a  probability map model respectively. \Cref{tab:results}~shows that the latter model has significantly lower perplexity, suggesting that the Gaussian model is somewhat too constrained. Qualitatively, \cref{fig:qualitative,fig:heat} show that both models make very reasonable predictions of uncertainty, with the uncertain area growing over time as expected.

\section{Conclusions}
\label{sec:conclusions}

In this paper we explored the possibility of using a single neural network for long-term prediction of mechanical phenomena. We considered in particular the problem of predicting the long-term motion of a cuboid sliding down a slope of unknown inclination and heterogeneous friction. Differently from many other approaches, we use the network {\em not} to predict some physical quantities to be integrated by a simulator, but to directly predict the complete trajectory of the object end-to-end.

Our results, obtained from extensive synthetic simulation, indicate that deep neural networks can successfully predict long-term trajectories without requiring explicit modeling of the underlying physics. They can also reliably estimate a distribution over such predictions to account for uncertainty in the data. Remarkably, these models are competitive with alternative predictors that have access to the ground-truth physical simulator, and outperform them when some of the physical parameters are not observable or known \emph{a-priori}. However, neural networks exhibit a limited capability to perform predictions outside the physical regimes observed during training. In other words, the internal representation of physics learned by such model is not as general as standard physical laws.

Several future directions remain to be explored. Given the accuracy of mechanical simulators, synthetic experiments are sufficient to assess the capability of networks to learn mechanical phenomena. However, the obvious next phase will be to test the framework on video footage obtained from real-world data in order to assess the ability to do so from visual data affected by real nuisance factors. The other important generalization is to consider more complex physical phenomena, including multiple sliding objects with possible interactions, rolling motion, and sliding over non-flat surfaces.

%\nocite{langley00}
%\bibliography{main}
\bibliographystyle{icml2017}

%\input{07_scratch}

%\clearpage
%\section{Appendix}
%\input{supp.tex}

\end{document}